\documentclass[10pt,twocolumn,letterpaper]{article}

\usepackage{hhline}
\usepackage{color, colortbl}
\usepackage{multirow}
\usepackage{multicol}
\usepackage{ijcb}
\usepackage{times}
\usepackage[pdftex]{graphicx}
\usepackage{amsmath}
\usepackage{amssymb}
\usepackage{subcaption}
\usepackage[utf8]{inputenc}
\usepackage{lipsum}  
\usepackage{stfloats}
\usepackage{booktabs}
\usepackage{xcolor}
\usepackage{makecell}
\usepackage{url}

\usepackage{authblk}
\usepackage{multirow}
\usepackage[overload]{empheq}

\ijcbfinalcopy 

\def\ijcbPaperID{112}

\ifijcbfinal\fi
\begin{document}

\title{Analysis of Manual and Automated Skin Tone Assignments for Face Recognition Applications}

\author[1]{K. S. Krishnapriya}
\author[1]{Michael C. King}
\author[2]{Kevin W. Bowyer}
\affil[1]{Florida Institute of Technology, Melbourne, Florida}
\affil[2]{University of Notre Dame, Notre Dame, Indiana}

\maketitle
\thispagestyle{empty}

\begin{abstract}
News reports have suggested that darker skin tone causes an increase in face recognition errors. The Fitzpatrick scale is widely used in dermatology to classify sensitivity to sun exposure and skin tone.
In this paper, we analyze a set of manual Fitzpatrick skin type assignments and also employ the individual typology angle to automatically estimate the skin tone from face images. The set of manual skin tone rating experiments shows that there are inconsistencies  between human  raters that are difficult to eliminate. Efforts to automate skin tone rating suggest that it is particularly challenging on images collected without a calibration object in the scene. However, after the color-correction, the level of agreement between automated and manual approaches is found to be 96\% or better for the MORPH images. 
To our knowledge, this is the first work to: (a) examine the consistency of manual skin tone ratings across observers, (b) document that there is substantial variation in the rating of the same image by different observers even when exemplar images are given for guidance and all images are color-corrected, and (c) compare manual versus automated skin tone ratings.
\end{abstract}

\section{Introduction}
Recent news articles have publicized the topic of bias in face recognition. For example, a BBC news article includes a figure with the caption ``Face recognition tech is less accurate the darker your skin tone'' \cite{BBC}, and a New York Times article states that ``... the darker the skin tone, the more errors  arise ...'' \cite{NYT}. While such articles are generally prompted by academic research results, we are not aware of any research that conclusively shows that darker skin tone is a primary causal factor in decreasing the accuracy of face recognition. 

In this paper, we report on an experiment to measure skin tone from face images in accordance with the Fitzpatrick skin type (FST) and the individual typology angle (ITA). The Fitzpatrick scale is a I (lighter) to VI (darker) rating that is widely used in dermatology \cite{Dermatology}. The Fitzpatrick scale has recently been adopted in various face recognition research studies \cite{gender_shades, ijbc, muthukumar2019color,Krishnapriya_TTS_2020}. Prior research \cite{del2006relationship, diversityinfaces} has also shown that skin tone assessment can be done directly from an image using automated individual typology angle measurement. Individual typology angle measurements are categorized into six skin type groups - very light, light, intermediate, tan, brown, and dark.

The major contributions of this work are as follows. One, we present the first analysis of the consistency of human rating of skin tones from images. The analysis of human ratings suggests that categorical labeling of skin tone by human observers is subjective, and there is some inconsistency across observers. We explore a series of improvements to encourage greater consistency of human ratings, initially adding color rectangles for reference, then exemplar images, then color-correction and exemplar images. Two, we customized an automated approach to skin tone assessment based on individual typology angle. This automated rating produces a level of agreement with manual ratings that is similar to the level of consistency between two human raters using the Fitzpatrick scale. The automated approach has obvious advantages in speed, scalability, cost, and consistency.
%------------------------------------------------------------------------
\section{Related Work}
Due to space limits, we give only a brief overview of how skin tone has been used in exploring accuracy differences in face image analysis. Initial research on recognition accuracy differences between African-American and Caucasian used race meta-data without taking individual skin tone into account. Perhaps the best known such study is Klare et al.~\cite{Klare_TIFS_2012}. They report that \cite{Klare_TIFS_2012}, ``the female, Black, and younger cohorts are more difficult to recognize for all matchers used in this study ...''. Krishnapriya et al.~\cite{Krishnapriya_CVPRW_2019} report that the impostor and genuine distributions for African-American are centered on higher similarity values than for Caucasian. This means that for a fixed decision threshold, African-Americans have a higher FMR and a lower FNMR. A recent NIST report \cite{FRVT_2019_Part3} finds that false matches are generally higher for West and East Africans and lower for Eastern Europeans, and also that with mugshot quality images, false non-matches are higher for Caucasians and lower for African-Americans. Wang et al.~\cite{wang2019racial} and Gong et al.~\cite{gong2019debface} also recently reported recognition accuracy results based on using race meta-data without skin tone ratings.

The Fitzpatrick scale \cite{Fitzpatrick_1988} is a I (lightest) to VI (darkest) rating of skin tone, used in dermatology to classify sensitivity to sun exposure. Skin tone, of course, varies among African-American individuals and among Caucasian individuals. Also, face morphology varies between the groups (and between gender) and between individuals, independent of skin tone variation. Lester et al. \cite{lester2020absence} reviewed the research literature on covid-19 skin manifestations in the context of the skin tone of subjects represented in research studies. A set of images from the literature were given Fitzpatrick skin tone categorization by a board-certified dermatologist \cite{lester2020absence} - ``... a board-certified dermatologist with expertise in diagnosing and treating patients with skin of colour (Fitzpatrick type IV-VI) evaluated each of the images and categorized them based on Fitzpatrick type I-VI.''  Lester et al. \cite{lester2020absence} also commented on the degree of uncertainty in these Fitzpatrick ratings - ``Our study is limited by the subjective assessment of skin type from a photograph. Lighting conditions including over-exposure may have made dark skin look lighter, and this may have led to some misclassification across one or two skin types. However, it is unlikely that lighting issues alone would result in skin types V or VI appearing as skin type I–III.'' This work underlines several important points. One, dermatological research on important current research questions is performed using a board-certified dermatologist's subjective assessment of Fitzpatrick skin type from photos. Whatever shortcomings subjective Fitzpatrick ratings from images have, there is not yet anything better to replace them. Two, the uncertainty due to varying illumination between images is acknowledged as possibly causing ``... some misclassification across one or two skin types'' \cite{lester2020absence}. Our experience with multiple observers rating the same set of controlled-acquisition face images is consistent with Lester et al. \cite{lester2020absence} on this point. And it makes sense that the more varied the illumination in a set of images, the larger the range of potential misclassification. In comparison to imagery in dermatology research publications, in-the-wild imagery used in face recognition research should be expected to have even larger misclassifications for Fitzpatrick skin type. In the context of our research in this paper, controlled-acquisition images such as those in MORPH will have less serious misclassification for Fitzpatrick skin type than would any of the in-the-wild datasets popular in face recognition research.

The use of Fitzpatrick ratings for face image analysis in the computer vision community appears to have started with the IARPA Janus dataset \cite{ijba} and with Buolamwini and Gebru's \cite{gender_shades} study. The Janus face image datasets \cite{ijbc} have meta-data for Fitzpatrick skin tone ratings obtained via Mechanical Turk. Lu et al.~\cite{Lu_TBIOM_2019} analyzed the Janus dataset and reported that recognition accuracy generally degraded with darker skin tone, but that skin tone VI had only the second-worst ROC. Buolamwini and Gebru \cite{gender_shades} generated skin tone ratings for images that they collected off the web and reported that each of the three gender classification tools studied was more accurate for lighter skin types than for darker. Muthukumar et al.~\cite{Muthukumar_2018} followed up with another study on gender classification tools and suggested that skin tone may not be the driving factor for accuracy differences. Krishnapriya et al.~\cite{Krishnapriya_TTS_2020} analyzed the distribution of skin tone ratings for images sampled from the center and from the high-similarity tail of the impostor distribution for African-American males. They reported that same-skin-tone image pairs occur more frequently in the high-similarity tail of the impostor distribution, but that darker skin tone does not appear to be a driving factor \cite{Krishnapriya_TTS_2020}.

Cook et al.~\cite{Cook_TBIOM_2019} and Howard et al.~\cite{Howard_BTAS_2019} analyzed recognition accuracy differences based on race meta-data and on a measure of skin reflectance. Exploiting the 18$\%$ gray background in a controlled enrollment image, they computed a measure of relative skin reflectance for each subject. They report that darker skin tone is associated with longer image acquisition times and with lower similarity scores for genuine image pairs, and that the skin reflectance measure was a better predictor than self-reported race labels. 

While various studies have used Fitzpatrick ratings assigned by viewers looking at images \cite{gender_shades, Howard_BTAS_2019, ijbc, muthukumar2019color, Muthukumar_2018}, there is little or no research on the consistency of the ratings.
%-------------------------------------------------------------------------
\section{Improvements in Manual Skin Tone Ratings}
We used the MORPH dataset \cite{morph_site} that contains mugshot-style images. MORPH was originally assembled and distributed to support research in face aging \cite{morph_paper}. The African-American male cohort of MORPH contains 36,838 images of 8,850 subjects in the curated version used in \cite{Krishnapriya_CVPRW_2019}. This cohort allows us to compute an impostor distribution in which we can analyze the effects of skin tone independent of factors of gender and race. The 8,850 African-American male subjects in MORPH are larger than the 3,531 total subjects in IJB-C. It is also far larger than the 562 (363+199) subjects in the dataset in \cite{Cook_TBIOM_2019}. Also, the dataset in \cite{Cook_TBIOM_2019}, in contrast to MORPH and IJB-C, is not available to the research community. Note that black rectangles are added over the eye regions to all the example images shown in this paper in an effort to respect individual anonymity and privacy.

For the analysis, we used an open-source deep CNN matcher called 
the ArcFace \cite{Deng_CVPR_2019}. One of the major advantages of ArcFace is that it optimizes the geodesic distance margin by utilizing the exact correspondence between arc and angle in a normalized hypersphere. The pre-trained ArcFace model used in this work can be found here: \url{https://github.com/deepinsight/insightface}

In this experiment, we sample 500 image pairs from each of two different regions, the center and high-similarity tail, of the African-American male impostor distribution. Each of these two sets of image pairs has just less than 1,000 unique images, as some individual images are repeated in multiple image pairs. There are 982 unique images of 915 persons from the center and 967 unique images of 872 persons from the high-similarity tail.

In our initial manual Fitzpatrick rating experiments, three different viewers independently examined each image to assign a Fitzpatrick score to the image. The three raters each assigned a Fitzpatrick score without knowing the region of the impostor distribution that an image came from and without knowing each others' ratings. If two or three of the viewers agreed on the skin tone rating, that was used as the rating for the image. If the three viewers gave different ratings, then the middle of the three ratings was used as the rating for the image. Fusing the results from different viewers is meant to be a possible improvement
over the ratings as used in \cite{gender_shades, muthukumar2019color}, avoiding anomalies specific to any one viewer.

For images from the center of the African-American male impostor distribution, all three reviewers gave the same rating to 352 images (36\%), two of three reviewers agreed for 586 images (60\%), and all three reviewers gave different ratings for 44 images (4\%). For images from the  high-similarity tail, all three reviewers gave the same rating to 302 images (31\%),  two of three reviewers agreed for 602 images (62\%), and all three reviewers gave different ratings for 63 images (7\%). The distributions of skin tone ratings by the three viewers are given on the top of Figure \ref{fig:Comparison_Plot}.

\subsection{Exemplar-guided Ratings}
\begin{figure}[!ht]
    \centering
    \includegraphics[width=0.45\textwidth]{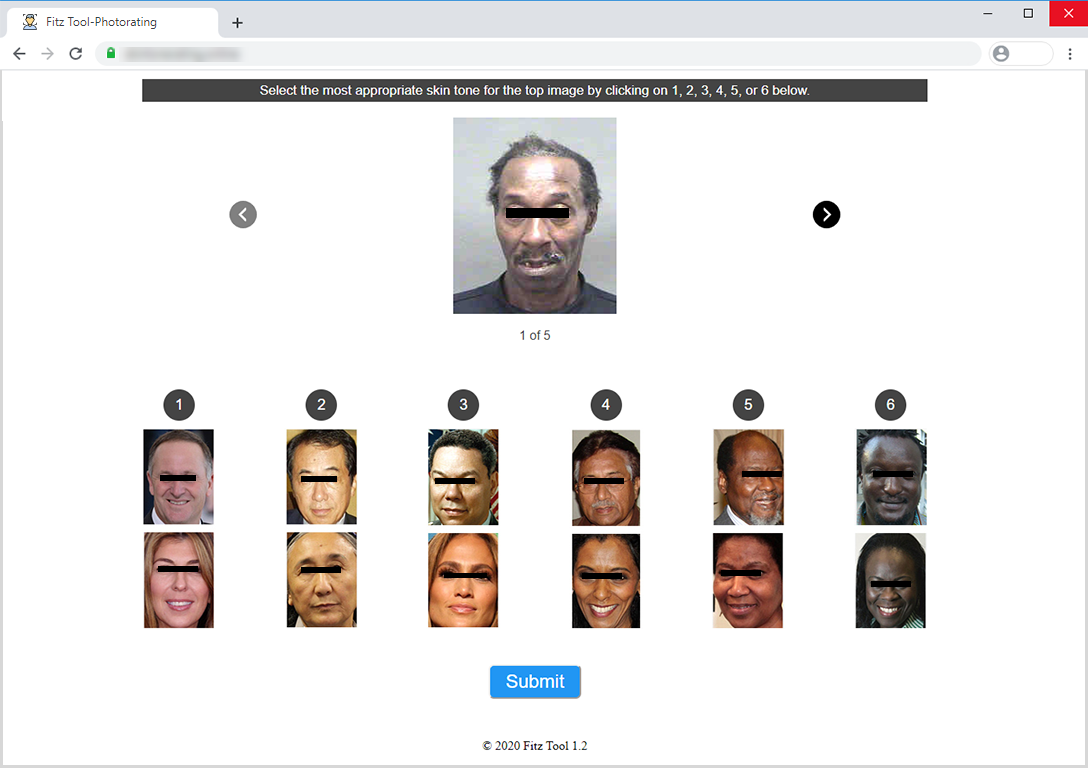}\hfill
    \caption{Snapshot of web-based skin tone rating tool showing exemplar images from IJB-C that correspond to six different Fitzpatrick ratings.}
    \label{fig:WebSkinToneTool}
\end{figure}

The initial experiment results show reasonable agreement between the three raters: e.g., agreement by at least 2 of 3 raters on 90\% of the images. Nevertheless, we refined the procedure for manual skin tone ratings in an effort to improve the consistency in the ratings. We posited that the rating task would give more consistent results if the raters had exemplar faces of different skin tones to compare with on each rating. Thus, we modified the software used to present images and record ratings to include exemplar images of the different Fitzpatrick ratings as shown in Figure \ref{fig:WebSkinToneTool}, along with the images to rate.

For exemplar images, we selected images from the well-known IJB-C dataset that has skin tone annotations in its metadata \cite{ijbc}. IJB-C has 3,531 subjects with 31,334 images and 117,542 frames from 11,779 videos \cite{ijbc}. From the metadata for IJB-C \cite{ijbc}, it appears to be a per-subject skin tone annotation. Based on a manual review of subjects in the IJB-C dataset, we chose six exemplars to use in our rating tool. Additionally, a web-based tool was developed to provide greater flexibility in getting viewers to rate images (see Figure \ref{fig:WebSkinToneTool}).  

We repeated the initial experiment, with the same images rated again by the same three raters using the version of the rating tool that displays exemplar images to encourage the use of consistent reference points. The comparison plots are given in Figure \ref{fig:Comparison_Plot}. The results suggest that the web-based tool with exemplar images does {\it not} cause greater consistency between the raters. Individual raters still seem to center their rating distribution differently on the FST. The other observation is that the ratings seem slightly more spread across 1 to 6 with the new exemplar-based tool compared to the baseline tool.

\subsection{Color-corrected Images and Exemplars}

MORPH images are acquired in a controlled environment with the subject standing in front of an 18\% gray background. This experiment is designed to normalize the face images so that the 18\% gray region is the same on average across all images. The motivating hypothesis is that varying color quality between images may contribute to inconsistent manual Fitzpatrick ratings.  

\begin{figure}[!ht]
    \centering
    \includegraphics[width=0.4\textwidth]{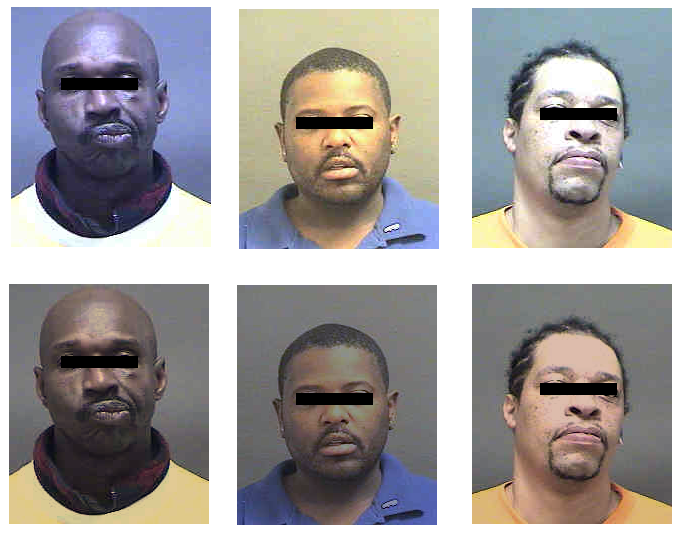}\hfill
    \caption{Images before color-correction (top) and images after color-correction (bottom).}
    \label{fig:color_correction_process}
\end{figure}

The 18\% gray is defined based on reflection, i.e., an 18\% gray surface reflects 18\% of the light that hits it \cite{middlegray}. The idea of 18\% gray in photography is to achieve middle gray to human perception. In different color spaces, middle gray may be defined differently. For example, in CIELAB, middle gray is defined to be 46.6\% brightness \cite{cielabgray} and in 24-bit color space, it is RGB (119, 119, 119) \cite{24gray}. This experiment has taken RGB (119, 119, 119) as the 18\% gray value \cite{24gray}. The steps followed in color-correction are:
\begin{itemize}
    \item Semantic segmentation of person and background in the given face image using a pre-trained model.
    \item Extract all (R, G, B) pixel values corresponds to the image background.
    \item Find the mean pixel value of Red, Green, and Blue components of the background. Let it be ($R_{avg}$, $G_{avg}$, $B_{avg}$). So, here,
    \\$R_{avg}$ = mean(all R components of the background)
    \\$G_{avg}$ = mean(all G components of the background)
    \\$B_{avg}$ = mean(all B components of the background)
    \item Find the color-correction factor from the background based on 18\% gray. It would be:
    \\$R_{const}$ = 119/$R_{avg}$
    \\$G_{const}$ = 119/$G_{avg}$
    \\$B_{const}$ = 119/$B_{avg}$
    \item Apply this color-correction factor to all the pixels in the original image. The corrected pixels would be:
    \\$R_{corrected}$ = $R_{const}$ * $R_{original}$
    \\$G_{corrected}$ = $G_{const}$ * $G_{original}$
    \\$B_{corrected}$ = $B_{const}$ * $B_{original}$
    \\and the pixels are not wrapped around 255.
\end{itemize}
We used a pre-trained model called DeepLab V3 \cite{deeplabv3} to segment the person and the background.  DeepLab has Xception \cite{xception} as its network backbone and was pre-trained on ImageNet \cite{imagenet_cvpr09}. Subjectively, this color-correction step  improves the visual quality of the original image (see Figure \ref{fig:color_correction_process}) and clearly makes the background more consistent across images.
\begin{figure}[!ht]
    \centering
    \includegraphics[width=0.45\textwidth]{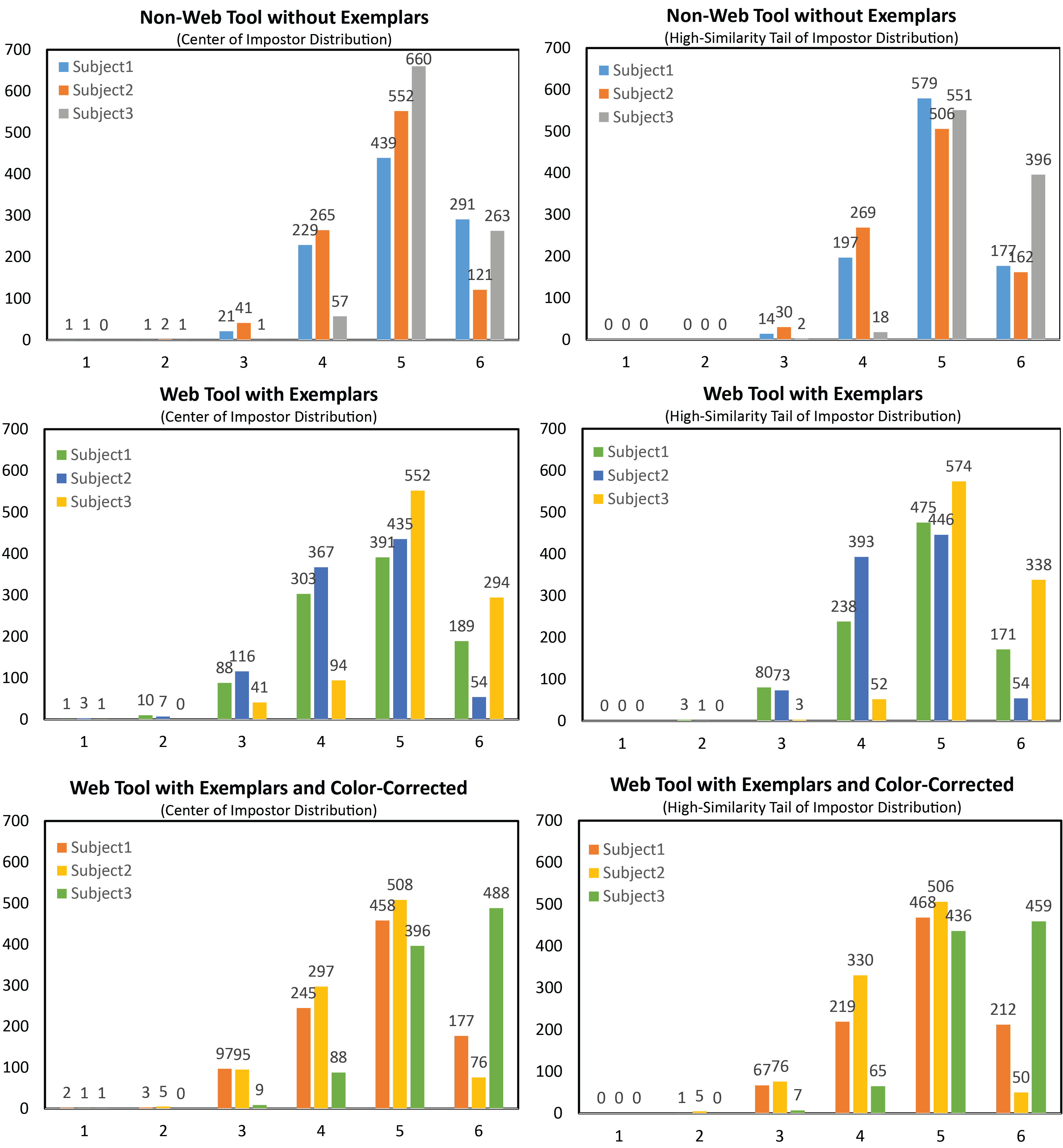}\hfill
    \caption{Comparison of the distribution of Fitzpatrick skin tone rating between individual raters with the baseline tool without exemplars (top), exemplar-guided on the raw images (middle) and exemplar-guided on the color-corrected images (bottom).}
    \label{fig:Comparison_Plot}
\end{figure}

\begin{figure}[!ht]
    \centering
    \includegraphics[width=1\columnwidth]{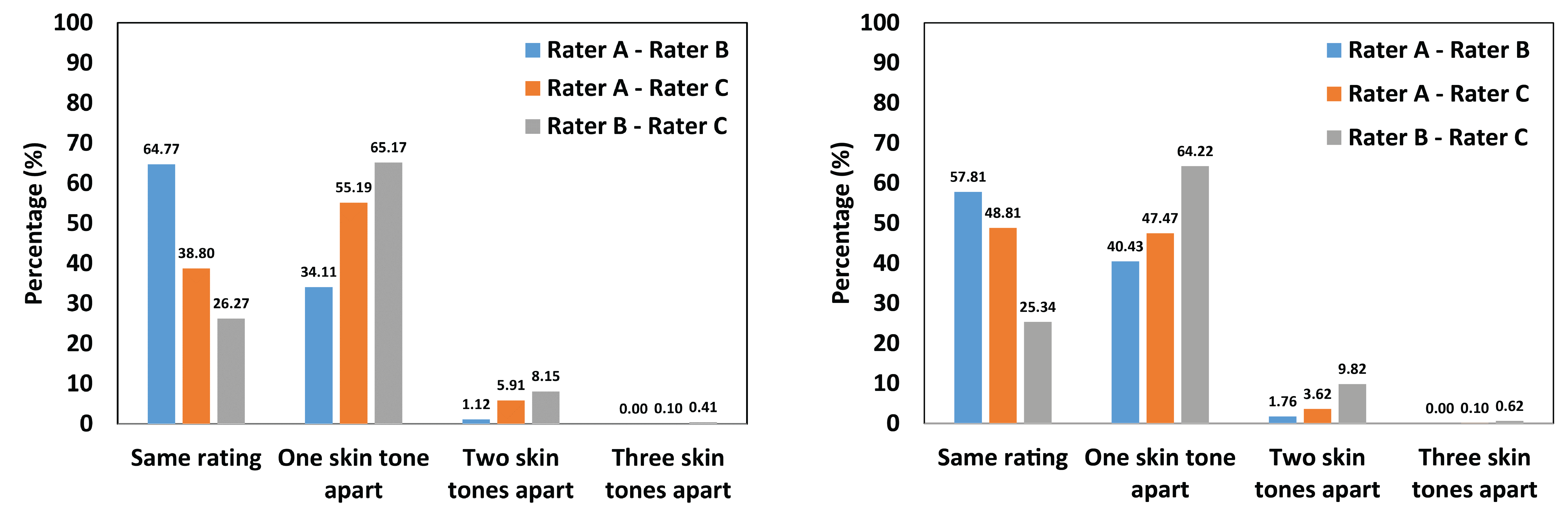}\hfill
    \caption{Inter-rater agreement for color-corrected images from the center (left) and high-similarity tail (right) of impostor distribution with exemplar-guided rating by the three viewers.}
    \label{fig:Inter-rater_Agreement_Morph}
    \end{figure}

We again repeated a set of ratings with the web-based tool, this time using the color-corrected versions of the original images and also having the exemplar images for reference. This set of comparison plots is given in Figure \ref{fig:Comparison_Plot}. The results suggest that color-correction gives an improved consistency, but the effect is small. The inter-rater agreement on the color-corrected images by three different viewers shows that it follows the same pattern whether the images are from the center of the distribution or the tail (see Figure \ref{fig:Inter-rater_Agreement_Morph}). Different pairs of raters have different levels of agreement. But allowing for a one rating level difference, all pairs have 89\% or better agreement.

Our sequence of manual skin tone rating experiments suggests that there is a level of inconsistency between human raters that is difficult to eliminate. Also, these experiments use relatively controlled images from the MORPH dataset. The inconsistencies observed here would likely be much greater, and the color-correction as implemented here would not be feasible for images from an in-the-wild dataset.

%-------------------------------------------------------------------------
\section{Automated Assignment of Skin Tones}

Our analysis of  the inconsistency in manual skin tone ratings motivates us to determine if automated skin tone ratings can be effectively used in face recognition research. Automated ratings would be 100\% consistent across two runs on the same image, repeatable across research groups, much faster and cheaper to acquire, and could be applied at a scale that is not feasible for manual ratings.

Prior research \cite{del2006relationship, diversityinfaces} has shown that skin tone assessment can be done directly from an image using computer-based individual typology angle (ITA) measurement. The ITA is calculated in the CIELab color space where $L$ represents the lightness, $a$ represents the chromaticity coordinate from green to red, and $b$ represents the chromaticity coordinate from blue to yellow. In this approach, we utilized ITA for representing the skin color \cite{chardon1991skin}, and it is calculated according to equation \ref{eqn:ITA}. ITA measurements are categorized into six skin type groups - very light (skin type I), light (skin type II), intermediate (skin type III), tan (skin type IV), brown (skin type V), and dark (skin type VI).

\subsection{Implementation Workflow}
\begin{figure}[!ht]
      \begin{subfigure}[b]{0.48\columnwidth}
          \begin{subfigure}[b]{1\columnwidth}
            \centering
            \includegraphics[width=\linewidth]{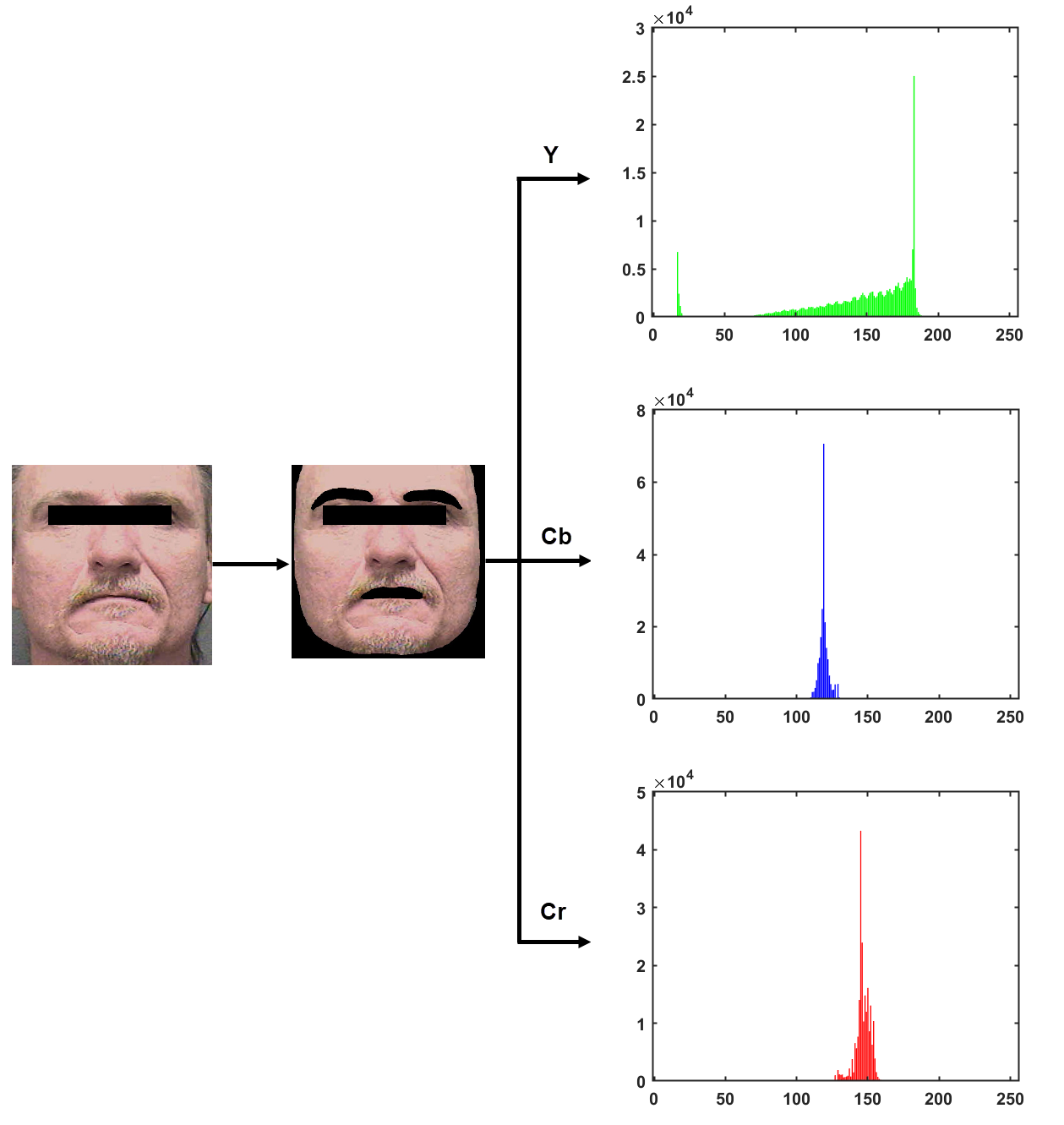}
          \end{subfigure}
          \caption{}
          \vspace{-0.5em}
           \label{fig:YCbCr_Plots_C}
      \end{subfigure}
      \hfill %%
      \begin{subfigure}[b]{0.48\columnwidth}
          \begin{subfigure}[b]{1\columnwidth}
            \centering
            \includegraphics[width=\linewidth]{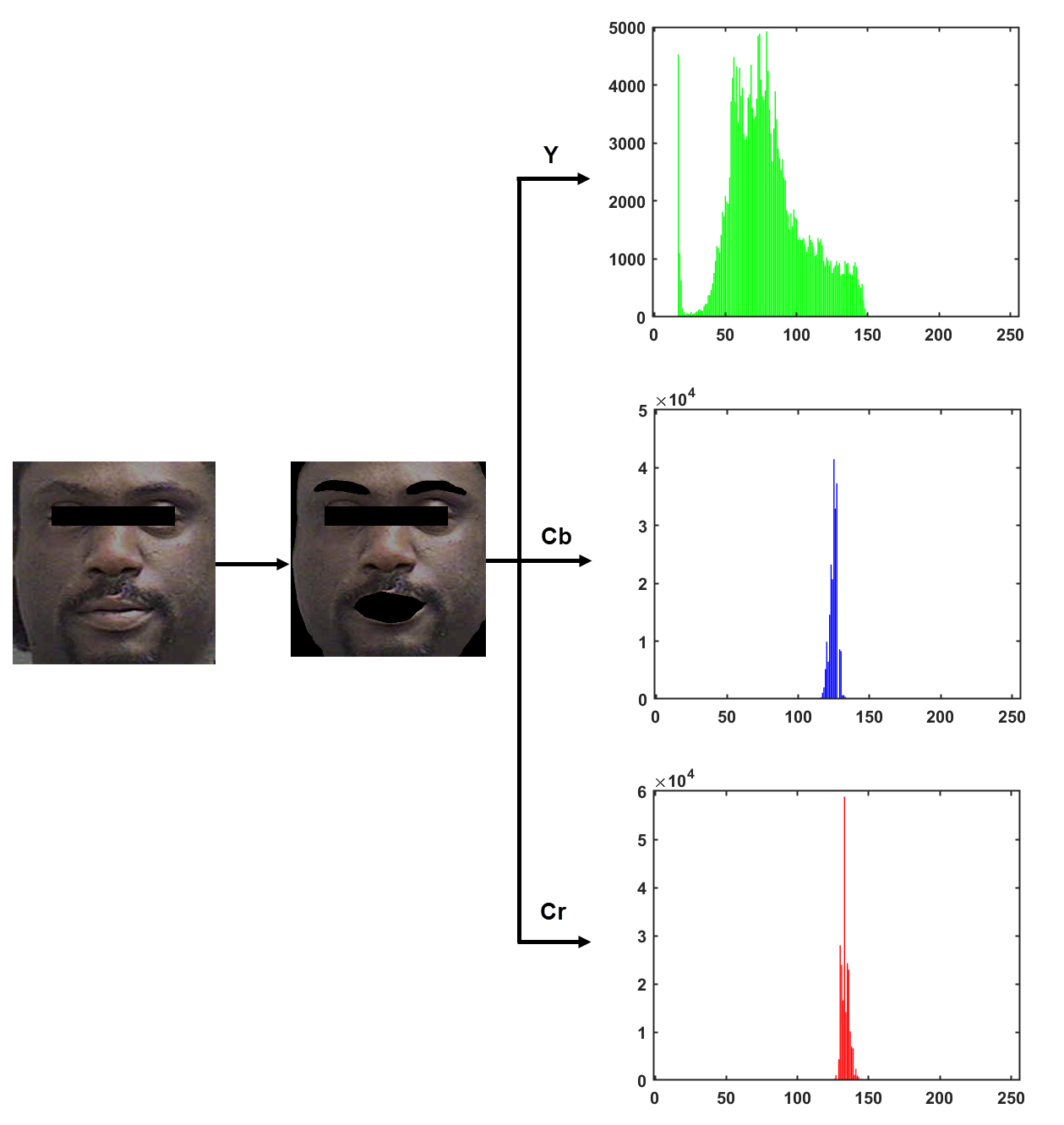}
          \end{subfigure}
          \caption{}
          \vspace{-0.5em}
          \label{fig:YCbCr_Plots_AA}
      \end{subfigure}
  \caption{Frequencies of Y, Cb, and Cr values across face skin pixels for (a) Caucasian (b) African-American.}
  \label{fig:YCbCr_Plots}
  \vspace{-1em}
\end{figure}

The selection of suitable color space for skin detection is an important factor in determining the higher probability of success. Prior research \cite{chai1999face,shaik2015comparative} has shown that the RGB color space is not usually preferred for color segmentation analysis because the brightness (luminance) component is not decoupled from the color information (chrominance). We can effectively utilize the chrominance information in $YCbCr$ color space for modeling the human skin color, and hence we propose thresholding on $YCbCr$ color space channels for skin detection. The $Y$ channel representing the brightness cannot be constrained here because when we are evaluating different datasets, and the images can be taken in different lighting conditions. Hence, with the $Y$ channel, it is difficult to determine if the variation in distribution are caused by different skin color or different lighting condition. In Figure \ref{fig:YCbCr_Plots}, we can see similar $Cr$ and $Cb$ distributions of skin color for Caucasian (see Figure \ref{fig:YCbCr_Plots_C}) and African-American (see Figure \ref{fig:YCbCr_Plots_AA}) images, and they do not seem to be affected by the variations in luminance. The evaluation of $Cb$ and $Cr$ channels across different sets of images showed that they are consistent across different demographic groups, and hence we can achieve better skin detection based on the thresholding on those two channels from an input image. The ranges mentioned for $Cb$ and $Cr$ in equation \ref{eqn:Ranges} were found to be the most suitable and representative of skin color for different sets of images we have tested.

\begin{figure}[!ht]
      \begin{subfigure}[b]{0.2\columnwidth}
          \begin{subfigure}[b]{1\columnwidth}
            \centering
            \includegraphics[width=\linewidth]{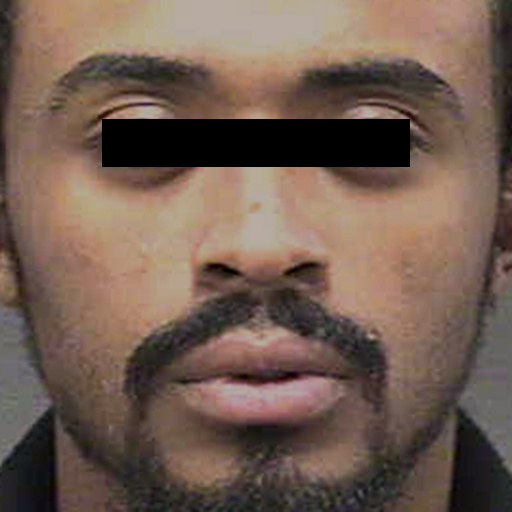}
          \end{subfigure}
          \caption{}
          \vspace{-0.5em}
           \label{fig:Dlib_preprocess}
      \end{subfigure}
      \hfill %%
      \begin{subfigure}[b]{0.2\columnwidth}
          \begin{subfigure}[b]{1\columnwidth}
            \centering
            \includegraphics[width=\linewidth]{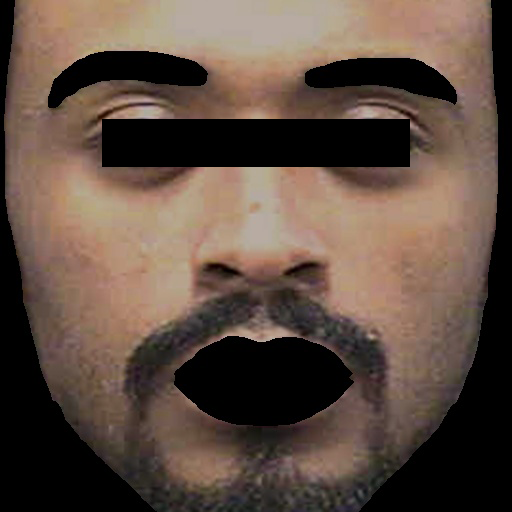}
          \end{subfigure}
          \caption{}
          \vspace{-0.5em}
          \label{fig:BiSeNet_skinextract}
      \end{subfigure}
      \hfill %%
      \begin{subfigure}[b]{0.2\columnwidth}
          \begin{subfigure}[b]{1\columnwidth}
            \centering
            \includegraphics[width=\linewidth]{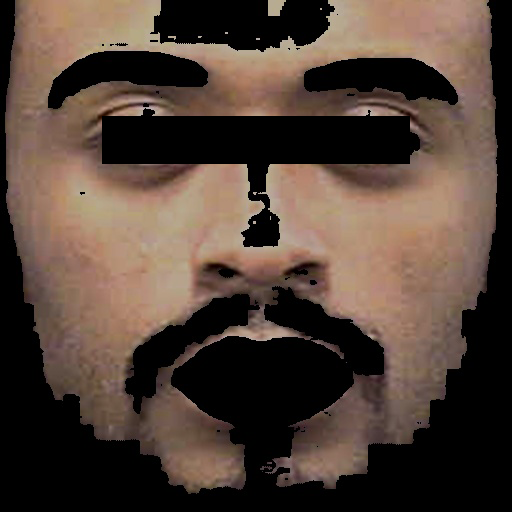}
          \end{subfigure}
          \caption{}
          \vspace{-0.5em}
          \label{fig:ycbcr_thresholding}
      \end{subfigure}
      \hfill %%
      \begin{subfigure}[b]{0.2\columnwidth}
          \begin{subfigure}[b]{1\columnwidth}
            \centering
            \includegraphics[width=\linewidth]{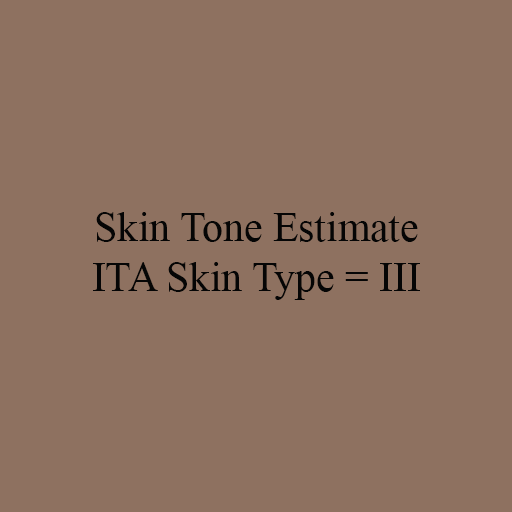}
          \end{subfigure}
          \caption{}
          \vspace{-0.5em}
          \label{fig:ita_count}
      \end{subfigure}
  \caption{Automated rating workflow (a) Dlib face detector (b) BiSeNet face skin segmentation (c) Cb and Cr thresholding (d) skin tone estimate.}
  \label{fig:automated_rating}
  \vspace{-1em}
\end{figure}

\begin{figure}[!ht]
    \centering
    \includegraphics[width=0.5\textwidth]{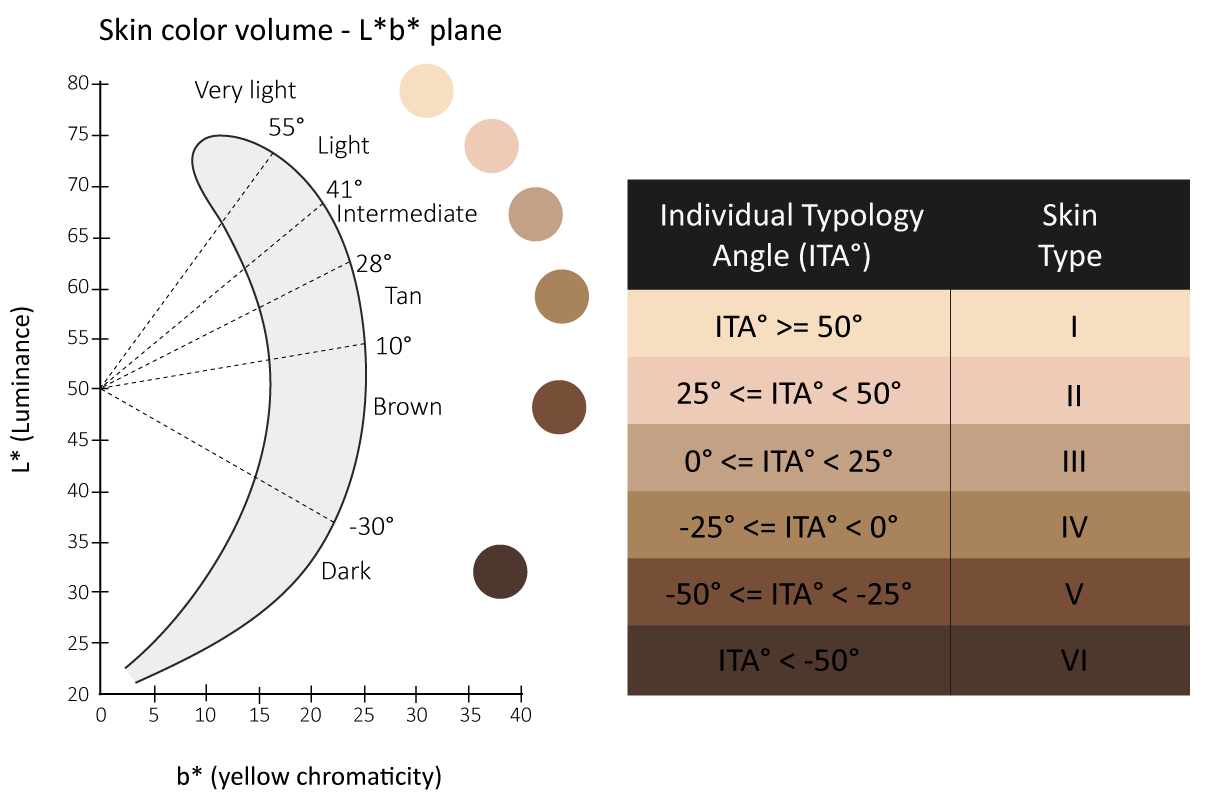}\hfill
    \caption{ITA skin type classification.}
    \label{fig:ITA_Skintype}
\end{figure}

This automated approach utilizes color-corrected images for skin tone assignments. Initially, face detection, alignment, and cropping are done using Dlib. We used a model called BiSeNet (Bilateral Segmentation Network) \cite{bisenet} for the face skin segmentation task.
BiSeNet was pre-trained on the CelebAMask-HQ \cite{CelebAMask-HQ} dataset that has 30,000 face images from CelebA \cite{celebA} and CelebA-HQ \cite{CelebA-HQ}. The eyes and lips regions are masked out intentionally to avoid any noise or occlusions like sunglasses while estimating the actual skin tone. The extracted face skin may contain over-exposed or under-exposed skin pixels due to illumination conditions. Thresholding on the $YCbCr$ color space is done to select the best skin pixels that are representative of a person's skin tone. Then from all the selected skin pixels, we find the mean pixel value for that face. The mean pixel value is converted to CIELab color space for its corresponding $L$ and $b$ values to compute the ITA. After finding the ITA measurement, we map that into skin types I to VI. The major steps in this automated skin tone rating include:
\begin{enumerate}
    \item Face detection, alignment, and crop using the Dlib face detector (see Figure \ref{fig:Dlib_preprocess}).
    \item Semantic segmentation of face skin from the detected face using the BiSeNet pre-trained model (see Figure \ref{fig:BiSeNet_skinextract}).
    \item Conversion to $YCbCr$ color space and applying the following thresholding on the $Cb$ and $Cr$ channels for skin pixels selection (see Figure \ref{fig:ycbcr_thresholding}).
    \begin{equation}\label{eqn:Ranges}
    \text{pixel}= 
    \begin{cases}
        \text{skin},& \text{if } 136 \leq $Cr$ \leq 173 \& 77 \leq $Cb$ \leq 127\\
        \text{non-skin},              & \text{otherwise}
    \end{cases}
    \end{equation}
    \item Calculate the mean pixel and compute its corresponding ITA using equation \ref{eqn:ITA}.
    \begin{equation}\label{eqn:ITA}
        ITA = \frac{arctan(\frac{(L - 50)}{b})*180}{\pi}
    \end{equation}
    \item Map the final ITA measurement to six skin type groups from very light (skin type I) to dark (skin type VI)(see Figure \ref{fig:ITA_Skintype}).
\end{enumerate}

\subsection{Manual vs. Automated Assignments}
To compare the consistency of manual consensus ratings to that of ITA ranges for skin type I to VI specified in \cite{chardon1991skin}, we analyzed 982 unique images from the center and 967 unique images from the high-similarity tail of the African-American male impostor distribution that have manual consensus ratings from three different viewers (exemplar-guided ratings on the color-corrected images). The mapping of these images on the ITA scale shows that there is substantial overlap within the ITA ranges. To give a reasonable means of assigning images to six categories based on ITA and inspired by the Fitzpatrick ratings, we examine custom threshold ranges for ITA skin type labels (customized ranges are shown in Figure \ref{fig:ITA_Skintype}). 

\begin{figure}[!ht]
    \centering
    \includegraphics[width=1\columnwidth]{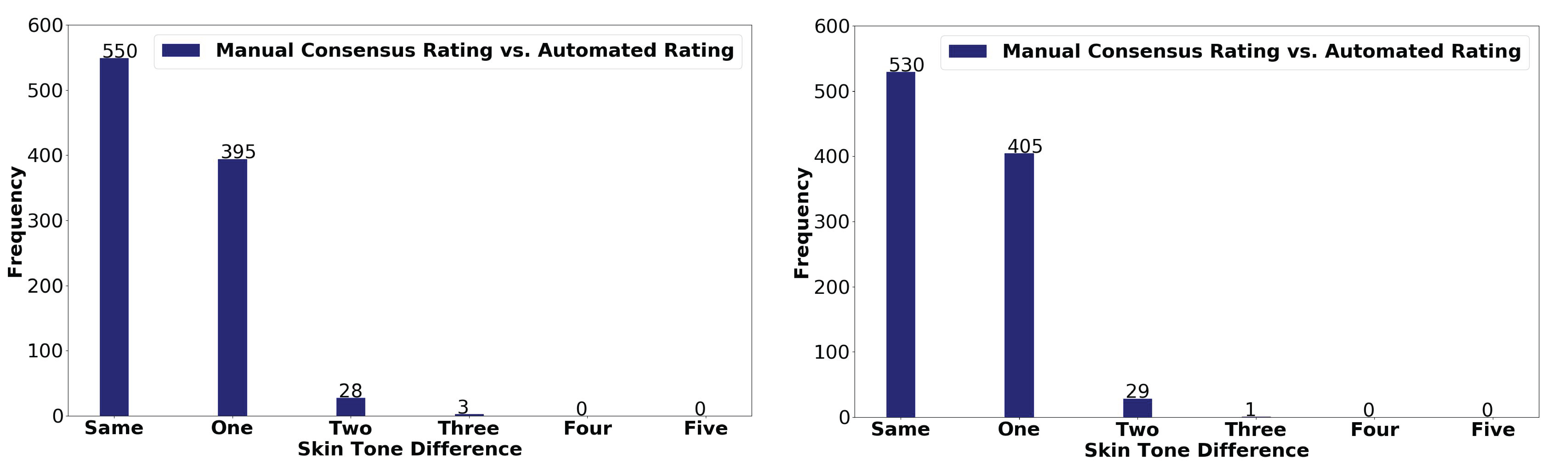}\hfill
    \caption{Distribution of difference in skin tone by the manual consensus rating and ITA automated approach for color-corrected images from the center (left) and high-similarity tail (right) of impostor distribution.}
    \label{fig:Morph_ManualvsAuto}
\end{figure}

There were six images from the center of the African-American male impostor distribution and two images from its high-similarity tail reported as failure-to-detect by Dlib. From the center, 550 images (56.4\%) had the same rating, 395 images had one-skin-tone-difference, 28 images had two-skin-tone-difference, and three images had more-than-two-skin-tone-difference with the automated approach and the manual consensus ratings. From its high-similarity tail, 530 images (54.9\%) had the same rating, 405 images had one-skin-tone-difference, 29 images had two-skin-tone-difference, and one image had more-than-two-skin-tone-difference with the automated approach and the manual consensus ratings. Considering the same or up to one skin tone of difference, there is 96\% or above consistency between the manual consensus ratings and automated ratings (see Figure \ref{fig:Morph_ManualvsAuto}). The manual-automated comparison based on the customized ITA ranges for MORPH images shows that the automated-to-consensus-manual consistency is as good as the consistency between any two individual raters. This is a good reason, along with the reproducibility, and the ease of use, to use the automated method.
%-------------------------------------------------------------------------
\section{Conclusion and Discussion}

This paper systematically analyzes approaches to estimate the skin tone of a person from an image. It describes methods for manual rating and proposes an automated skin tone assignment approach for greater ease of use, scalability, and reproducibility. We also discuss the pros and cons of each approach. The major conclusions and takeaways from these experiments are as follows.

The categorical labeling of skin tone by human observers can be subjective and inconsistent. The same images were observed to have been rated differently by different raters or different at different times by the same rater. Several studies mention that it is subjective even by trained practitioners \cite{borza2018automatic, arigbabu2015recent}. Inter-rater agreement for the manual rating by three different viewers shows that different pairs of persons have different levels of agreement. But allowing for a one rating level difference, all pairs have 89\% or better (See Figure \ref{fig:Inter-rater_Agreement_Morph}). 

\begin{figure}[!ht]
    \centering
    \includegraphics[width=0.5\textwidth]{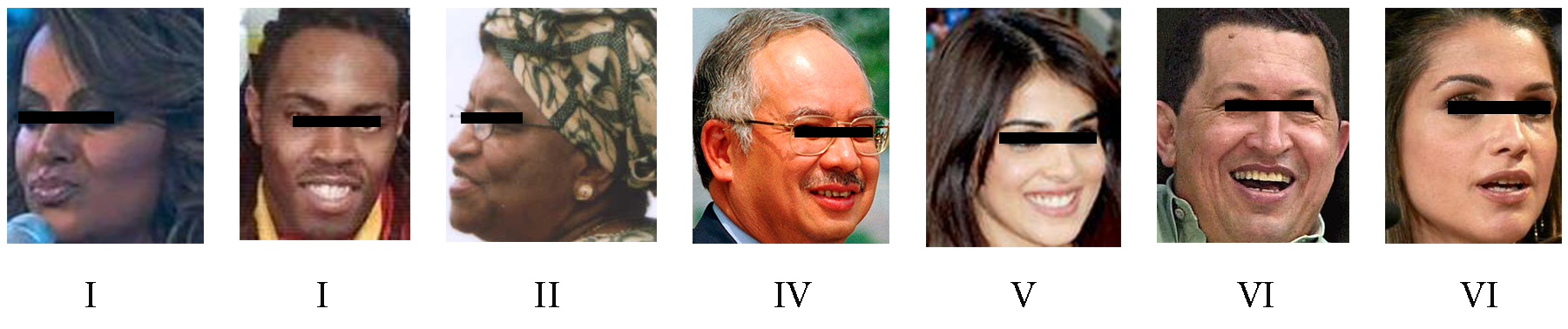}\hfill
    \caption{Example misclassifications found in the IJB-C dataset. The skin tone annotations given in the metadata for these images from left to right are I, I, II, IV, V, VI, and VI respectively.}
    \label{IJBC_Fitzpatrick_SkinTone}
    \end{figure}

\begin{figure}[!ht]
    \centering
    \includegraphics[width=0.5\textwidth]{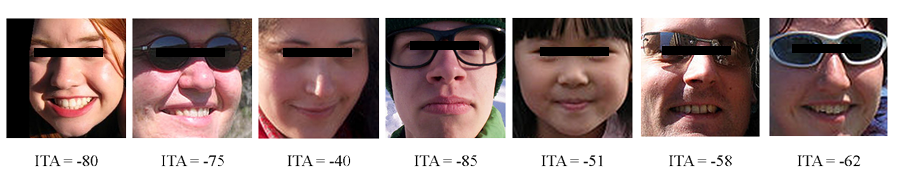}\hfill
    \caption{Example misclassifications found in the DiF dataset. The ITA skin tone annotations given in the metadata for these images from left to right are -80, -75, -40, -85, -51, -58, and -62 respectively, and all those maps to ITA skin type "dark".}
    \label{DiF_Fitzpatrick_SkinTones}
\end{figure}

While skin tone rating from a color image may seem simple in concept, it is complex and quite challenging. Prior research has been conducted on re-purposed images with skin tone rating assessed by humans in accordance with the Fitzpatrick scale \cite{FDA,Fitzpatrick_1988}, and more recently by computer-based individual typology angle measurement \cite{diversityinfaces}. Both of these techniques are flawed in that the human rating (for example, see Figure \ref{IJBC_Fitzpatrick_SkinTone}) and automated rating (for example, see Figure \ref{DiF_Fitzpatrick_SkinTones}) are often on non-ideal images that have been taken in non-controlled environments.

The effort to automate the process of Fitzpatrick skin tone labeling from images is quite challenging. The analysis of ITA distributions for images consistently rated as I to VI with manual consensus rating shows that we can do retrospective labeling of images using the Fitzpatrick scale as a guide. The results, however, will be inherently noisy. We have not found strong evidence to support that this noise due to the experience level of the raters but more likely can be attributed to the range of colors for skin-related pixels that are induced by varying lighting conditions and sensor characteristics. The automated-to-consensus-manual consistency is as  good  as  the  consistency between two individual raters. The automated skin tone rating algorithm was developed and tested using mugshot-style images. The efficacy of this approach on in-the-wild face images is unknown and will need to be evaluated in future work.
%-------------------------------------------------------------------------
{\small
\bibliographystyle{ieee}
\bibliography{egbib.bib}

\begin{thebibliography}{10}\itemsep=-1pt

\bibitem{24gray}
Cie color calculator.
\newblock {\em brucelindbloom}.
\newblock \url{http://www.brucelindbloom.com/index.html?ColorCalculator.html}.

\bibitem{morph_site}
Morph dataset.
\newblock \url{https://www.faceaginggroup.com/morph}.

\bibitem{middlegray}
J.~Aldred.
\newblock What is middle grey and why does it even matter?, 2018.
\newblock
  \url{https://www.diyphotography.net/what-is-middle-grey-and-why-does-it \\
  even-matter/}.

\bibitem{arigbabu2015recent}
O.~A. Arigbabu, S.~M.~S. Ahmad, W.~A.~W. Adnan, and S.~Yussof.
\newblock Recent advances in facial soft biometrics.
\newblock {\em The Visual Computer}, 31(5):513--525, 2015.

\bibitem{Dermatology}
O.~Arosarena.
\newblock Options and challenges for facial rejuvenation in patients with
  higher fitzpatrick skin phototypes.
\newblock {\em JAMA Facial Plastic Surgery}, 2015.

\bibitem{borza2018automatic}
D.~Borza, A.~S. Darabant, and R.~Danescu.
\newblock Automatic skin tone extraction for visagism applications.
\newblock In {\em VISIGRAPP (4: VISAPP)}, pages 466--473, 2018.

\bibitem{gender_shades}
J.~Buolamwini and T.~Gebru.
\newblock Gender shades: Intersectional accuracy disparities in commercial
  gender classification.
\newblock In {\em Proceedings of Machine Learning Research 81: Conference on
  Fairness, Accountability, and Transparency}, 2018.

\bibitem{chai1999face}
D.~Chai and K.~N. Ngan.
\newblock Face segmentation using skin-color map in videophone applications.
\newblock {\em IEEE Transactions on circuits and systems for video technology},
  9(4):551--564, 1999.

\bibitem{chardon1991skin}
A.~Chardon, I.~Cretois, and C.~Hourseau.
\newblock Skin colour typology and suntanning pathways.
\newblock {\em International journal of cosmetic science}, 13(4):191--208,
  1991.

\bibitem{deeplabv3}
L.-C. Chen, G.~Papandreou, F.~Schroff, and H.~Adam.
\newblock Rethinking atrous convolution for semantic image segmentation.
\newblock {\em arXiv preprint arXiv:1706.05587}, 2017.

\bibitem{xception}
F.~Chollet.
\newblock Xception: Deep learning with depthwise separable convolutions.
\newblock In {\em CVPR 2017}.

\bibitem{Cook_TBIOM_2019}
C.~M. Cook, J.~J. Howard, Y.~B. Sirotin, J.~L. Tipton, and A.~R. Vemury.
\newblock Demographic effects in facial recognition and their dependence on
  image acquisition: An evaluation of eleven commercial systems.
\newblock {\em IEEE Transactions on Biometrics, Behavior, and Identity
  Science}, 40(1), 2019.

\bibitem{del2006relationship}
S.~Del~Bino, J.~Sok, E.~Bessac, and F.~Bernerd.
\newblock Relationship between skin response to ultraviolet exposure and skin
  color type.
\newblock {\em Pigment cell research}, 19(6):606--614, 2006.

\bibitem{imagenet_cvpr09}
J.~Deng, W.~Dong, R.~Socher, L.-J. Li, K.~Li, and L.~Fei-Fei.
\newblock {ImageNet: A Large-Scale Hierarchical Image Database}.
\newblock In {\em CVPR09}, 2009.

\bibitem{Deng_CVPR_2019}
J.~Deng, J.~Guo, N.~Xue, and S.~Zafeiriou.
\newblock Arcface: Additive angular margin loss for deep face recognition.
\newblock In {\em IEEE Conference on Computer Vision and Pattern Recognition},
  2019.

\bibitem{Fitzpatrick_1988}
T.~B. Fitzpatrick.
\newblock The validity and practicality of sun-reactive skin types i through
  vi.
\newblock {\em Archives of Dermatology}, 124(6):869–871, 1988.

\bibitem{FDA}
U.~Food and D.~Administration.
\newblock Your skin.
\newblock \url{https://www.fda.gov/radiation-emitting-products/
  tanning/your-skin}{, (last accessed July 2020)}.

\bibitem{cielabgray}
S.~Geffert.
\newblock Adopting iso standards for museum imaging.
\newblock {\em imagingetc.com, Inc}, 2008.

\bibitem{gong2019debface}
S.~Gong, X.~Liu, and A.~K. Jain.
\newblock Debface: De-biasing face recognition.
\newblock {\em arXiv preprint arXiv:1911.08080}, 2019.

\bibitem{FRVT_2019_Part3}
P.~Grother, M.~Ngan, and K.~Hanaoka.
\newblock Nistir 8280: Ongoing face recognition vendor test (frvt) part 3:
  Demographic effects.
\newblock \url{https://nvlpubs.nist.gov/nistpubs/ir/2019/NIST.IR.8280.pdf}.

\bibitem{Howard_BTAS_2019}
J.~J. Howard, Y.~B. Sirotin, and A.~Vemury.
\newblock The effect of broad and specific demographic homogeneity on the
  imposter distributions and false match rates in face recognition algorithm
  performance.
\newblock In {\em 10th IEEE International Conference on Biometrics: Theory,
  Applications and Systems}, September 2019.

\bibitem{CelebA-HQ}
T.~Karras, T.~Aila, S.~Laine, and J.~Lehtinen.
\newblock Progressive growing of gans for improved quality, stability, and
  variation.
\newblock {\em arXiv preprint arXiv:1710.10196}, 2017.

\bibitem{Klare_TIFS_2012}
B.~F. Klare, M.~J. Burge, J.~C. Klontz, R.~W. {Vorder Bruegge}, and A.~K. Jain.
\newblock Face recognition performance: Role of demographic information.
\newblock {\em IEEE Transactions on Information Forensics and Security},
  7(6):1789--1801, 2012.

\bibitem{ijba}
B.~F. Klare, B.~Klein, E.~Taborsky, A.~Blanton, J.~Cheney, K.~Allen,
  P.~Grother, A.~Mah, and A.~K. Jain.
\newblock Pushing the frontiers of unconstrained face detection and
  recognition: Iarpa janus benchmark a.
\newblock In {\em CVPR 2015n}.

\bibitem{Krishnapriya_TTS_2020}
K.~Krishnapriya, V.~Albiero, K.~Vangara, M.~C. King, and K.~Bowyer.
\newblock Issues related to face recognition accuracy varying based on race and
  skin tone.
\newblock {\em IEEE Transactions on Technology and Society}, 1(1), 2020.

\bibitem{Krishnapriya_CVPRW_2019}
K.~Krishnapriya, K.~Vangara, M.~C. King, V.~Albiero, and K.~Bowyer.
\newblock Characterizing the variability in face recognition accuracy relative
  to race.
\newblock In {\em Computer Vision and Pattern Recognition (CVPR) Workshops},
  2019.

\bibitem{CelebAMask-HQ}
C.-H. Lee, Z.~Liu, L.~Wu, and P.~Luo.
\newblock Maskgan: Towards diverse and interactive facial image manipulation.
\newblock In {\em CVPR 2020}.

\bibitem{lester2020absence}
J.~Lester, J.~Jia, L.~Zhang, G.~Okoye, and E.~Linos.
\newblock Absence of images of skin of colour in publications of covid-19 skin
  manifestations.
\newblock {\em British Journal of Dermatology}, 183(3):593--595, 2020.

\bibitem{celebA}
Z.~Liu, P.~Luo, X.~Wang, and X.~Tang.
\newblock Deep learning face attributes in the wild.
\newblock In {\em ICCV 2015}.

\bibitem{NYT}
S.~Lohr.
\newblock Facial recognition is accurate, if you’re a white guy.
\newblock {\em The New York Times}, 9 February 2018.
\newblock
  https://www.nytimes.com/2018/02/09/technology/facial-recognition-race-artificial-intelligence.html.

\bibitem{Lu_TBIOM_2019}
B.~Lu, J.~Chen, C.~D. Castillo, and R.~Chellappa.
\newblock An experimental evaluation of covariates effects on unconstrained
  face verification.
\newblock {\em IEEE Transactions on Biometrics, Behavior, and Identity
  Science}, 40(1), 2019.

\bibitem{ijbc}
B.~Maze, J.~Adams, J.~A. Duncan, N.~Kalka, T.~Miller, C.~Otto, A.~K. Jain,
  W.~T. Niggel, J.~Anderson, J.~Cheney, et~al.
\newblock Iarpa janus benchmark - c: Face dataset and protocol.
\newblock In {\em 2018 International Conference on Biometrics (ICB)}, pages
  158--165. IEEE, 2018.

\bibitem{diversityinfaces}
M.~Merler, N.~Ratha, R.~S. Feris, and J.~R. Smith.
\newblock Diversity in faces.
\newblock {\em arXiv preprint arXiv:1901.10436}, 2019.

\bibitem{muthukumar2019color}
V.~Muthukumar.
\newblock Color-theoretic experiments to understand unequal gender
  classification accuracy from face images.
\newblock In {\em Proceedings of the IEEE Conference on Computer Vision and
  Pattern Recognition Workshops}, 2019.

\bibitem{Muthukumar_2018}
V.~Muthukumar, T.~Pedapati, N.~Ratha, P.~Sattigeri, C.-W. Wu, B.~Kingsbury,
  A.~Kumar, S.~Thomas, A.~Mojsilovic, and K.~R. Varshney.
\newblock Understanding unequal gender classification accuracy from face
  images.
\newblock In {\em https://arxiv.org/abs/1812.00099}, 2018.

\bibitem{morph_paper}
K.~Ricanek and T.~Tesafaye.
\newblock {MORPH: A longitudinal image database of normal adult
  age-progression}.
\newblock In {\em International Conference on Automatic Face and Gesture
  Recognition}, 2006.

\bibitem{shaik2015comparative}
K.~B. Shaik, P.~Ganesan, V.~Kalist, B.~Sathish, and J.~M.~M. Jenitha.
\newblock Comparative study of skin color detection and segmentation in hsv and
  ycbcr color space.
\newblock {\em Procedia Computer Science}, 57(12):41--48, 2015.

\bibitem{BBC}
M.~Wall.
\newblock Biased and wrong? facial recognition tech in the dock.
\newblock {\em BBC News}, 8 July 2019.
\newblock https://www.bbc.com/news/business-48842750.

\bibitem{wang2019racial}
M.~Wang, W.~Deng, J.~Hu, X.~Tao, and Y.~Huang.
\newblock Racial faces in the wild: Reducing racial bias by information
  maximization adaptation network.
\newblock In {\em ICCV 2019}.

\bibitem{bisenet}
C.~Yu, J.~Wang, C.~Peng, C.~Gao, G.~Yu, and N.~Sang.
\newblock Bisenet: Bilateral segmentation network for real-time semantic
  segmentation.
\newblock In {\em Proceedings of the European conference on computer vision
  (ECCV)}, pages 325--341, 2018.

\end{thebibliography}
}

\end{document}